\documentclass[runningheads]{llncs}
\usepackage{graphicx}
\usepackage[dvipsnames]{xcolor}
\usepackage{subcaption}

\usepackage{amsmath}
\usepackage{amsfonts}
\usepackage{multirow}
\usepackage{booktabs}
\usepackage{cite}

\begin{document} 
\title{Data AUDIT: Identifying \underline{A}ttribute \underline{U}tility- and \underline{D}etectability-\underline{I}nduced Bias in \underline{T}ask Models}
%
\titlerunning{Data AUDIT}
%


\author{Mitchell~Pavlak\thanks{Equal contribution}\inst{1,2} \and
Nathan~Drenkow*\inst{1} \and
Nicholas~Petrick\inst{2} \and
Mohammad~Mehdi~Farhangi\inst{2} \and 
Mathias~Unberath\inst{1}}

\authorrunning{M. Pavlak, N. Drenkow et al.}

%
\institute{The Johns Hopkins University, Baltimore MD, USA \and
Center for Devices and Radiological Health, U.S. Food and Drug Administration, Silver Spring MD, USA}

\maketitle              

\begin{abstract}
To safely deploy deep learning-based computer vision models for computer-aided detection and diagnosis, we must ensure that they are robust and reliable. 
Towards that goal, algorithmic auditing has received substantial attention. To guide their audit procedures, existing methods rely on heuristic approaches or high-level objectives (e.g., non-discrimination in regards to protected attributes, such as sex, gender, or race).
However, algorithms may show bias with respect to various attributes beyond the more obvious ones, and integrity issues related to these more subtle attributes can have serious consequences. To enable the generation of actionable, data-driven hypotheses which identify specific dataset attributes likely to induce model bias, we contribute a first technique for the rigorous, quantitative screening of medical image \textit{datasets}. Drawing from literature in the causal inference and information theory domains, our procedure decomposes the risks associated with dataset attributes in terms of their detectability and utility (defined as the amount of information knowing the attribute gives about a task label). To demonstrate the effectiveness and sensitivity of our method, we develop a variety of datasets with synthetically inserted artifacts with different degrees of association to the target label that allow evaluation of inherited model biases via comparison of performance against true counterfactual examples. Using these datasets and results from hundreds of trained models, we show our screening method reliably identifies nearly imperceptible bias-inducing artifacts. Lastly, we apply our method to the natural attributes of a popular skin-lesion dataset and demonstrate its success. Our approach provides a means to perform more systematic algorithmic audits and guide future data collection efforts in pursuit of safer and more reliable models.

\keywords{Bias, shortcut learning, fairness, algorithmic auditing, datasets}
\end{abstract}

\section{Introduction}
Continual advancement of deep learning algorithms for medical image analysis has increased the potential for their adoption at scale.  Across a wide range of medical applications including skin lesion classification~\cite{esteva2017dermatologist, soenksen2021using}, detection of diabetic retinopathy in fundus images~\cite{gulshan2016development}, detection of large vessel occlusions in CT~\cite{Murray2020Artificial}, and detection of pneumonia in chest x-ray~\cite{rajpurkar2017chexnet}, deep learning algorithms have pushed the boundaries close to or beyond human performance. 

However, with these innovations has come increased scrutiny of the integrity of these models in safety critical applications. Prior work~\cite{jabbour2020deep, degrave2021ai, geirhos2020shortcut} has found that deep neural networks are capable of exploiting spurious features and other shortcuts in the data which are not causally linked to the task of interest such as using dermascopic rulers as cues to predict melanoma~\cite{winkler2019association, winkler2021association, bevan2021skin} or associating the presence of a chest drain with pneumothorax in chest X-ray analysis~\cite{oakden2020hidden}.  The exploitation of such shortcuts by DNNs may have serious bias/fairness implications~\cite{Gichoya2022ReadingRace, Glocker2022Multitask} and negative ramifications for model generalization~\cite{degrave2021ai,oakden2020hidden}.

As attention to these issues grows, recent legislation has been proposed that would require the algorithmic auditing and impact assessment of ML-based automated decision systems \cite{Wyden2022AccountabilityAct}. However,  without clearly defined strategies for selecting attributes to audit for bias,  impact assessments risk being constrained to only legally protected categories and may miss more subtle shortcuts and data flaws that prevent the achievement of important model goals\cite{Raji2022Fallacy,Sambasivan2021DataCascades}.
Our goal in this work is to develop objective methods for generating data-driven hypotheses about the relative level of risk of various attributes to better support the efficient, comprehensive auditing of any model trained on the same data.   

Our method generates targeted hypotheses for model audits by assessing (1) how feasible it is for a downstream model to detect and exploit the presence of a given attribute from the image alone (detectability), and (2) how much information the model would gain about the task labels if said attribute were known (utility).  
Causally irrelevant attributes with high utility and detectability become top priorities when performing downstream model audits. We demonstrate high utility complicates attempts to draw conclusions about the detectability of attributes and show our approach succeeds where unconditioned approaches fail.
We rigorously validate our approach using a range of synthetic artifacts which allow us to expedite the auditing of models via the use of true counterfactuals. We then apply our method to a popular skin lesion dataset where we identify a previously unreported potential shortcut.

\section{Related Work}

Issues of bias and fairness are of increasing concern in the research community. Recent works such as~\cite{Seyyed-Kalantari2021-aj, Seyyed-Kalantari2021-fa, Glocker2022-st} identify cases where trained DNNs exhibit performance disparities across protected groups for chest x-ray classification tasks. Of interest to this work,~\cite{o2022evaluating} used Mutual Information-based analysis to examine the robustness of DNNs on dermascopy data and observed performance disparities with respect to typical populations of interest (i.e., age, sex) as well as less commonly audited dataset properties (e.g., image hue, saturation). In addition to observed biased performance in task models,~\cite{Glocker2022Multitask, Gichoya2022ReadingRace} show that patient race (and potentially other protected attributes) may be implicitly encoded in representations extracted by DNNs on chest x-ray images.  A more general methodology for performing algorithmic audits in medical imaging is also proposed in~\cite{Liu2022-zm}. In contrast to our work, these methods focus on individual, biased task models without considering the extent to which those biases are induced by the causal structure of the training/evaluation data.

In addition to model auditing methods, a number of metrics have been proposed to quantify bias. A recent study~\cite{Aka2021-kq} compared several and recommend normalized pointwise mutual information due to its ability to measure associations in the data while accounting for chance.  Also relevant to this work, ~\cite{Henry_Hinnefeld2018-vi} provides an analysis of fairness metrics and guidelines for metric selection in the presence of dataset bias. However, these studies focus primarily on biases identifiable through dataset attributes alone and do not consider whether those attributes are detectable in the image data itself.  

Lastly,~\cite{Sambasivan2021DataCascades} found pervasive \textit{data cascades} where data quality issues compound and cause downstream adverse impacts for vulnerable groups.  However, their study was qualitative and no methods for automated dataset auditing were introduced. Bissoto et al.~\cite{bissoto2020debiasing, Bissoto2019-ek} consider the impact of bias in dermatological data by manipulating images to remove potential causally-relevant features while measuring a model's ability to still perform the lesion classification task. Closest to our work,~\cite{Reimers2021-vi} takes a causal approach to shortcut identification by using conditional dependence tests to determine whether DNNs rely on specific dataset attributes for their predictions. In contrast, our work focuses on screening \textit{datasets} for attributes that induce bias in task models. As a result, we directly predict attributes to act as a strong upper bound on detectability and use normalized, chance-adjusted dependence measures to obtain interpretable metrics that we show correlate well with the performance of task models.

\section{Methods}
\begin{figure}[t!]
    \centering
    \includegraphics[width=0.6\textwidth]{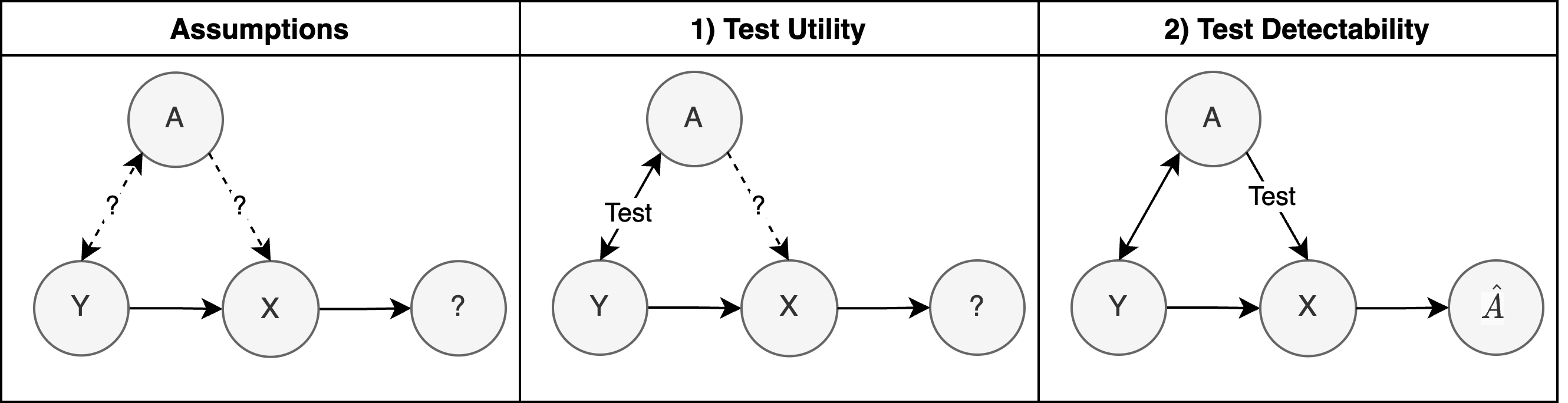}
    \caption{Relationships assessed in the attribute screening protocol. DNNs are trained to predict $\hat{A}$ from X for use in estimating attribute detectability. 
    }
    \label{fig:audit-procedure}
\end{figure}

To audit at the dataset level, we perform a form of causal discovery to identify likely relationships between the task labels, dataset attributes represented as image metadata, and features of the images themselves (as illustrated in Figure~\ref{fig:audit-procedure}).

We start from a set of labels $\{Y\}$, attributes $\{A\}$, and images $\{X\}$.  We assume that $Y$ (the diagnosis) is the causal parent of $X$ (the image) given that the diagnosis affects the image appearance but not vice versa~\cite{castro2020causality}. Then the dataset auditing procedure aims to assess the existence and relative strengths of the following two relationships: (1) \textit{Utility}: $A \leftrightarrow Y$ and (2) \textit{Detectability}: $A \rightarrow X$.  The utility measures whether a given attribute shares any relationship with the label. The presence of this relationship for $A$ that are not clinically relevant (e.g., sensor type or settings) represents increased potential for biased outcomes. However, not every such attribute carries the same risk for algorithmic bias. Crucially, relationship (2) relates to the detectability of the attribute itself. If our test for (2) finds the existence of relationship $A \rightarrow X$ is probable, we consider the attribute detectable. Dataset attributes identified as having positive utility with respect to the label (1) and detectable in the image (2) are classified as potential shortcuts and pose the greatest risk to models trained on this dataset.

\subsubsection{Causal Discovery with Mutual Information}
Considering attributes in isolation, we assess attribute utility and detectability from an information theoretic perspective. In particular, we recognize first that the presence of a relationship between $A$ and $Y$ can be measured via their Mutual Information: $MI(A; Y) = H(Y) - H(Y|A)$. 
$MI$ measures the information gained (or reduction in uncertainty) about $Y$ by observing $A$ (or vice versa) and $MI(A;Y)=0$ occurs when $A$ and $Y$ are independent.  We rely on the faithfulness assumption which implies that a causal relationship exists between $A$ and $Y$ when $MI(A;Y) > 0$. From an auditing perspective, we aim to identify the presence and relative magnitude of the relationship but not necessarily the nature of it.

Attributes identified as having a relationship with $Y$ are then assessed for their detectability (i.e., condition (2)). We determine detectability by training a DNN on the data for predicting each attribute.  Because we wish to audit the entire dataset for bias, we cannot rely on a single train/val/test split.  Instead, we partition the dataset into $k$ folds (typically 3) and finetune a sufficiently expressive DNN on the train split of each fold
to predict the given attribute $A$. We then generate unbiased predictions for the entire dataset by taking the output $\hat{A}$ from each DNN evaluated on their respective test split. We measure the Conditional Mutual Information over all predictions: $CMI(\hat{A}; A|Y) =  H(\hat{A}|Y) - H(\hat{A}|A,Y)$. 
$CMI(A,\hat{A}|Y)$ measures information shared between attribute $A$ and its prediction $\hat{A}$ when controlling for information provided by $Y$.  Since relationship $A$ and $Y$ was established via $MI(A;Y)$, we condition on label $Y$ to understand the extent to which attribute $A$ can be predicted from images when accounting for features associated with $Y$ that may also improve the prediction of $A$. Similar to $MI$, $CMI(\hat{A};A|Y) > 0$ implies $A\rightarrow \hat{A}$ exists.

To determine independence and account for bias and dataset specific effects, we include permutation based shuffle tests from \cite{Runge2018GeneralCausal,runge2018NNPermutation}. These approaches replace values of $A$ with close neighbors to approximate the null hypothesis that the given variables are conditionally independent. By calculating the percentile of $CMI(A; \hat{A}|Y)$ among all $CMI(A_\pi; \hat{A}| Y)$ (where $A_\pi$ are permutations of $A$), we estimate the probability our samples are independent while adjusting for estimator bias and dataset specific effects. 
To make raw CMI and MI statistics interpretable for magnitude based comparison between attributes, we include adjustments for underlying distribution entropy and chance assignment as per \cite{vinh2010ami, Aka2021-kq} (See supplementary materials).

\section{Experiments and Discussion}
To demonstrate the effectiveness of our method, we first conduct a series of experiments using synthetically altered skin lesion data from the HAM10000 dataset where we precisely create, control, and assess biases in the dataset. After establishing the accuracy and sensitivity of our method on synthetic data, we apply our method to the natural attributes of HAM10000 in Experiment 5 (\ref{experiment_5}).

\begin{figure}[t!]
    \centering
    \includegraphics[width=0.7\linewidth]{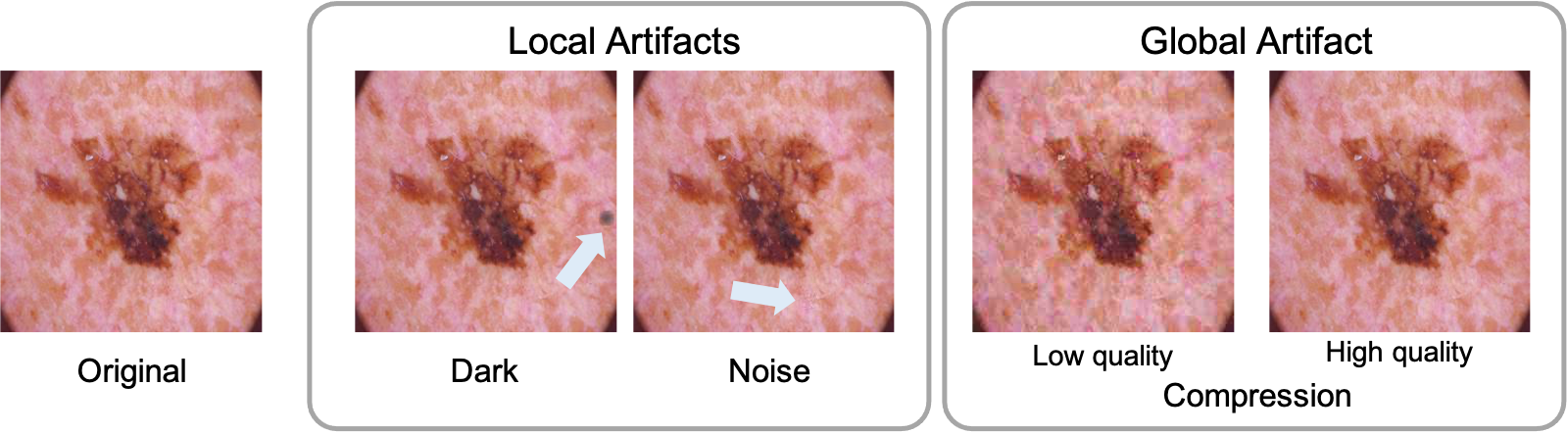}
    \caption{Examples of synthetic artifacts with varying image effects. 
    }
    \label{fig:examples}
\end{figure}

\textbf{Datasets: }
We use publicly available skin lesion data from the HAM10000~\cite{tschandl2018ham10000} dataset with additional public metadata from \cite{bevan2021skin}. The dataset consists of 10,015 dermascopic images collected from two sites and after repeated images of the same lesion we are left with 7,387 images. The original dataset has seven diagnostic categories: we focus on predicting lesion malignancy as a challenging and practical task. For our synthetic trials, we perturb this dataset with a variety of realistic artifacts (e.g., Figure~\ref{fig:examples}), and vary association with the malignant target label. We use this to generate datasets with attributes that have known utility and detectability as well as ground truth counterfactuals for task model evaluation. Further details in supplementary materials.

\textbf{Training Protocol: }
For attribute prediction networks used by our detectability procedure, we finetune ResNet18~\cite{he2016deep} models with limited data augmentation. For the malignancy prediction task we use Swin Transformer~\cite{liu2021swin} tiny models with RandAugment augmentation to show detectability results generalize to stronger architectures. All models were trained using balanced sampling with a batch size of 128 and the AdamW optimizer with a learning rate of 5e-5, linear decay schedule, and default weight decay and momentum parameters. For each trial, we use three fold cross validation and subdivide each training fold in a (90:10) train:validation split to select the best models for the relevant test fold. By following this procedure, we get unbiased artifact predictions over the entire dataset for use by MI estimators.  We generally measure model performance via the Receiver Operating Characteristic Area Under the Curve (AUC).

 \begin{figure}[t!]
     \centering
     \begin{subfigure}[b]{0.4\textwidth}
         \centering
         \includegraphics[height=4.5cm]{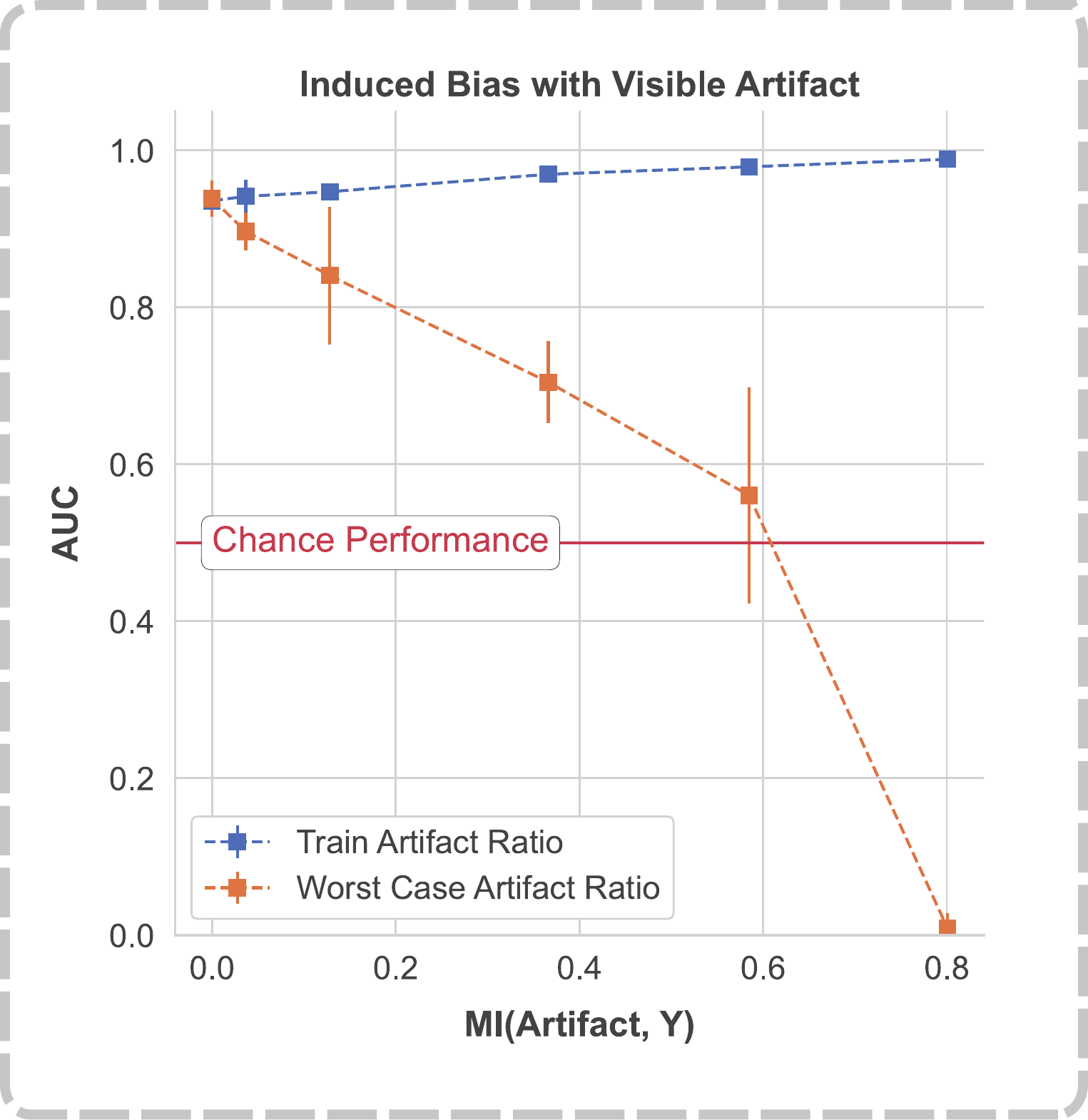}
         \caption{ }
         \label{fig:Induced_Bias_Visible}
     \end{subfigure}
          \begin{subfigure}[b]{0.4\textwidth}
         \centering
         \includegraphics[height=4.5cm]{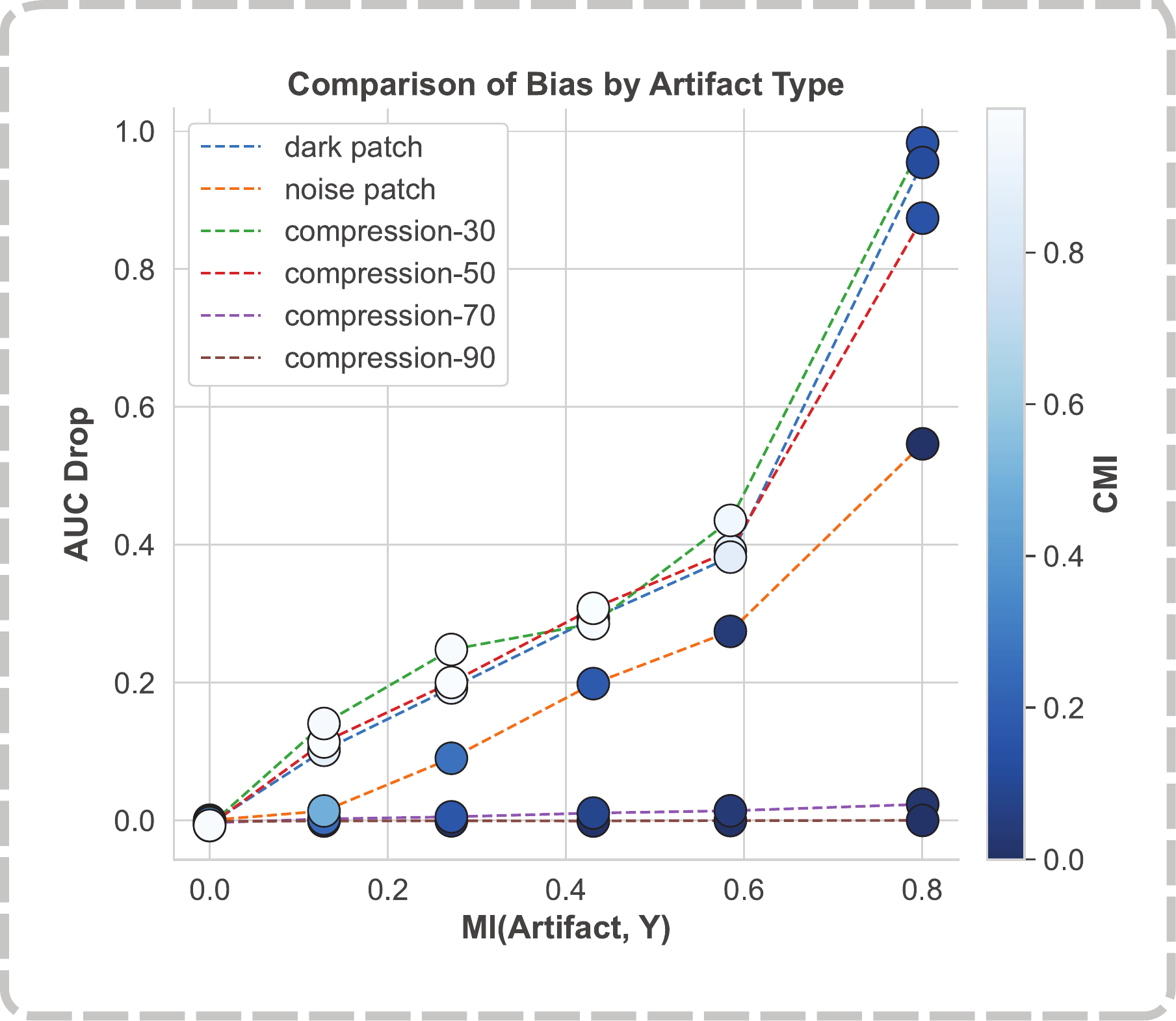}
         \caption{ }
         \label{fig:combined_auc_cami_ami}
     \end{subfigure}
     
    \caption{(a) Performance of task models trained on data with known detectable artifacts introduced with various positive correlations to the malignant class and evaluated on our worst-case counterfactual set. (b) Performance drop of models trained with artifacts of unknown detectability. For each,  $MI(A,Y)$, $CMI(A,\hat{A}|Y)$ values are estimated empirically with normalization and adjustment for chance applied. 
    } 
    \label{fig:bias_by_mi}
\end{figure}

 \subsection{Experiment 1: Induced Bias versus Relationship Strength} \label{experiment_1}

For this trial, we select an artifact that \textit{we are certain is visible} (JPEG compression at quality 30 applied to 1000 images), and seek to understand how the relationship between attribute and task label influences the task model's reliance on the attribute. The artifact is introduced with increasing utility such that the probability of the artifact is higher for cases that are malignant. Then, we create a worst case counterfactual set, where each malignant case does not have the artifact, and each benign case does. In Figure \ref{fig:Induced_Bias_Visible}, we see performance rapidly declines \textit{below random chance} as utility increases. 

\subsection{Experiment 2: Detectability of Known Invisible Artifacts} \label{experiment_2}

In the previous section, we showed that the utility $A\leftrightarrow Y$ directly impacts the task model bias, \textit{given A is visible in images}. However, it is not always obvious whether an attribute is visible. In \textit{Reading Race}, Gichoya et al. showed racial identity can be predicted with high AUC from medical images where this information is not expected to be preserved. 
Here, we show CMI represents a promising method for determining attribute detectability while controlling for attribute information communicated through labels and not through images.

\begin{figure}[h!]
     \centering
         \includegraphics[width=\textwidth]{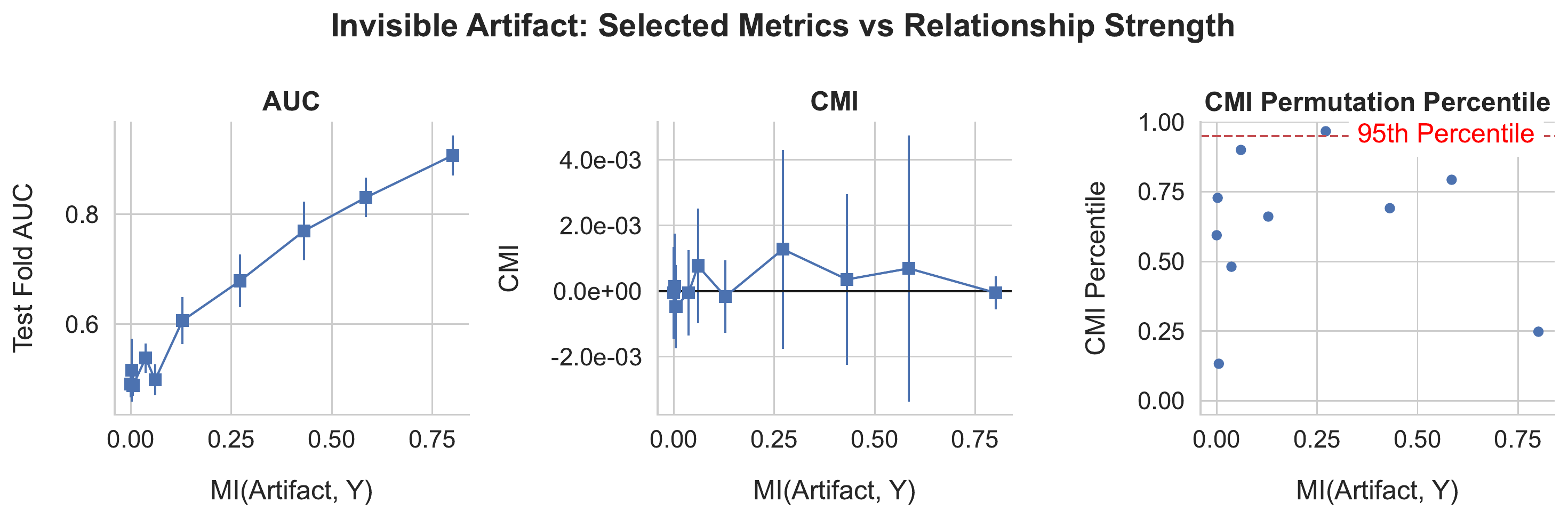}
        \caption{Left: AUC vs utility MI(A;Y). The models learn to fit a non-existent artifact given sufficient utility $A\leftrightarrow Y$. Middle:  CMI reported with 95\% CIs (calculated via bootstrap) for the same models and predictions, each interval includes zero, implying independence. Right: Percentile of CMI statistic vs CMI computed from 1000 samples permuted to be conditionally independent.}
        \label{fig:invisible_artifact}
\end{figure}

Specifically, we consider the case of an "invisible artifact". We make no changes to images, but instead create a set of randomized labels for our non-existent artifact that have varying correlation with the task labels ($A\leftrightarrow Y$).  As seen in Figure \ref{fig:invisible_artifact}, among cases where the invisible artifact and task label have a reasonable association, models tasked with predicting the invisible artifact perform well above random chance, seemingly indicating that these artifacts are visible in images. However, by removing the influence of task label and related image features by calculating $MI(A,\hat{A}|Y)$, we clearly see that the artifact predictions are independent of the labels, meaning there is no visible attribute; instead, all information about the attribute is inferred from the task label in this case.

\subsection{Experiment 3: Conditioned Detectability versus Ground Truth}\label{experiment_3}

To verify that the conditional independence testing procedure does not substantially reduce our ability to correctly identify artifacts that truly are visible, we introduce Gaussian noise with standard deviation decreasing past human perceptible levels. Now, because we introduce artifacts with no relationship to detectable image characteristics, AUC represents a valid ground truth. As can be seen from Table \ref{tab:noise_detect_vs_gt}, the performance drop from conditioning is minimal.

\begin{table}[t!]
\centering
\caption{Detectability of Gaussian noise with varying strength vs independence testing-based percentile and CMI (*normalized and adjusted for chance).}\label{tab:noise_detect_vs_gt}
\resizebox{0.7\linewidth}{!}{%
{\setlength{\tabcolsep}{4pt}
\renewcommand{\arraystretch}{1.25}
\begin{tabular}{|l|lllllllll|}
\hline
 \multicolumn{10}{|c|}
 {\bfseries{Noise Detectability vs Ground Truth}}
 \\
 \hline
\multirow{2}{*}{\textbf{Metric}} & \multicolumn{9}{c|}{\textbf{Gaussian Noise $\sigma$}}                                                             \\ 
\cline{2-10}
                                 & \multicolumn{1}{l|}{\bfseries{.5}}        & \multicolumn{1}{l|}{\bfseries{.4}}        & \multicolumn{1}{l|}{\bfseries{.3}}        & \multicolumn{1}{l|}{\bfseries{.2}}        & \multicolumn{1}{l|}{\bfseries{.1}}        & \multicolumn{1}{l|}{\bfseries{.05}}       & \multicolumn{1}{l|}{\bfseries{.01}}           & \multicolumn{1}{l|}{\bfseries{.001}}          & \bfseries{0}             \\ \hline
\bfseries{AUC}                              & \multicolumn{1}{l|}{1.0$\pm$0} & \multicolumn{1}{l|}{1.0$\pm$0} & \multicolumn{1}{l|}{1.0$\pm$0} & \multicolumn{1}{l|}{1.0$\pm$0} & \multicolumn{1}{l|}{1.0$\pm$0} & \multicolumn{1}{l|}{1.0$\pm$0} & \multicolumn{1}{l|}{0.71$\pm$.075} & \multicolumn{1}{l|}{0.53$\pm$.017} & 0.52$\pm$.036 \\ \hline
\bfseries{CMI*}                              & \multicolumn{1}{l|}{1.0}       & \multicolumn{1}{l|}{1.0}       & \multicolumn{1}{l|}{1.0}       & \multicolumn{1}{l|}{1.0}       & \multicolumn{1}{l|}{0.997}     & \multicolumn{1}{l|}{0.997}     & \multicolumn{1}{l|}{0.0496}        & \multicolumn{1}{l|}{-4.44e-4}      & -4.7e-5       \\ \hline
\bfseries{Percentile}                       & \multicolumn{1}{l|}{1.0}       & \multicolumn{1}{l|}{1.0}       & \multicolumn{1}{l|}{1.0}       & \multicolumn{1}{l|}{1.0}       & \multicolumn{1}{l|}{1.0}       & \multicolumn{1}{l|}{1.0}       & \multicolumn{1}{l|}{1.0}           & \multicolumn{1}{l|}{0.089}         & 0.595         \\ 
\hline
\end{tabular}%
}
}
\end{table}

\subsection{Experiment 4: Relationship and Detectability vs Induced Bias }\label{experiment_4}
Next, we consider how utility and detectability together relate to bias. We introduce a variety of synthetic artifacts and levels of bias and measure the drop in AUC that occurs when evaluated on a test set with artifacts introduced in the same ratio as training versus the worst case ratio as defined in Experiment 1. 

Of 36 unique attribute-bias combinations trialed, 32/36 were correctly classified as visible via permutation test with 95\% cutoff percentile. The remaining four cases were all compression at quality 90, and had negligible impact on task models (mean drop in AUC of $-0.0003\pm.0006$). Inspecting results from Figure \ref{fig:combined_auc_cami_ami}, we see the relative strength of the utility, $A\leftrightarrow Y$, correlates with the AUC drop observed. This implies utility represents a useful initial metric to predict the risk of an attribute. The detectability ($CMI$), decreases as utility ($MI$) increases, implying the two are not independent. However, detectability among attributes with equal utility for each level above 0 have statistically significant correlations with drops in AUC (Kendall's $\tau$ of 0.800, 0.745, 0.786, 0.786, 0.716 respectively). From this, we expect that for attributes with roughly equal utility, the more detectable attributes are more likely to result in biased task models.

\subsection{Experiment 5: HAM10000 Natural Attributes}\label{experiment_5}
\begin{figure}[t!]
     \centering
         \includegraphics[width=.57\textwidth]{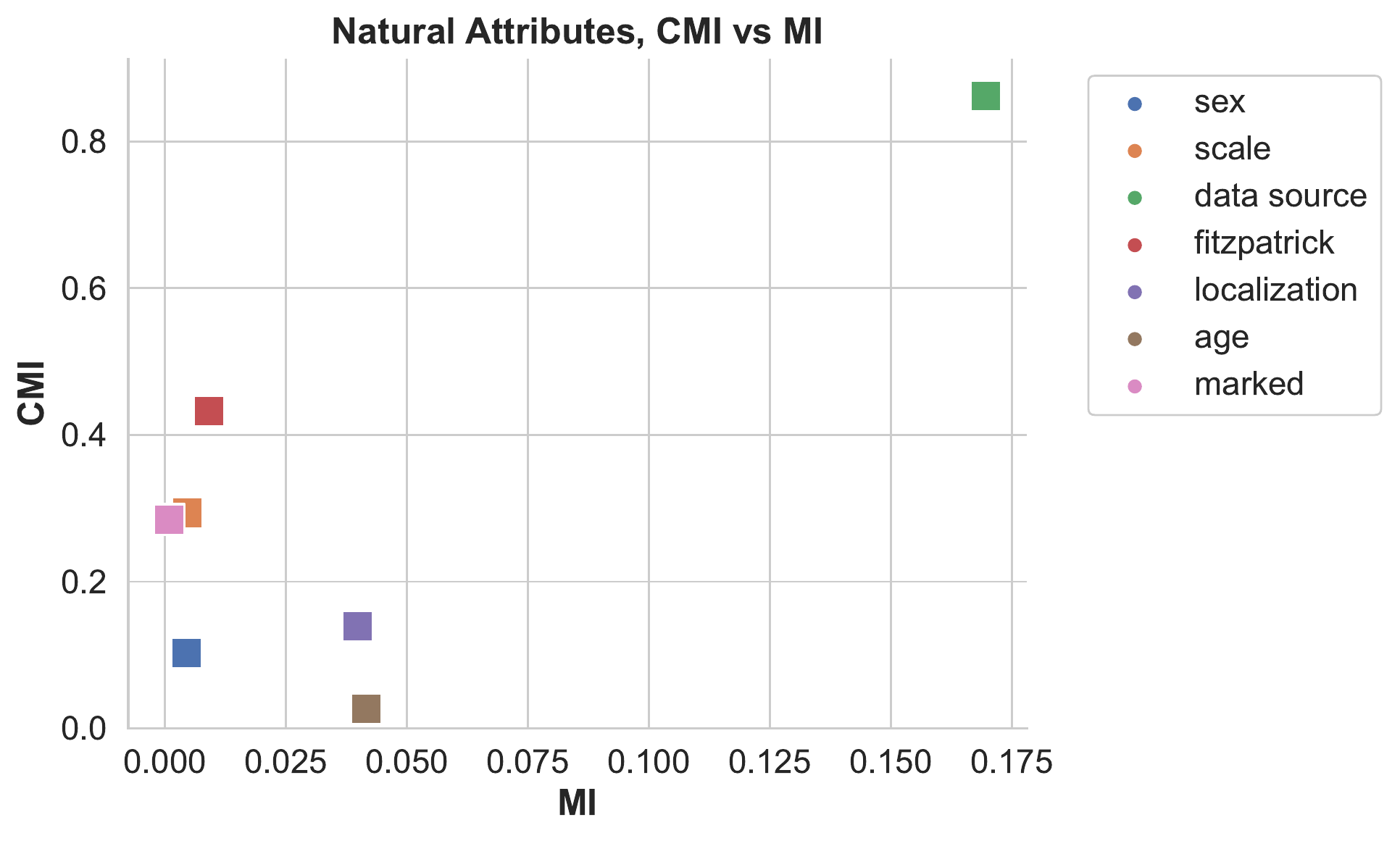}
        \caption{Detectability vs. utility for natural attributes in HAM10000.  Attributes with high $CMI$ and $MI$ which are non-causally related to the diagnosis pose the greatest risk.}
        \label{fig:natural_attributes}
\end{figure}

Last, we run our screening procedure over the natural attributes of HAM10000 and find that all pass conditional independence tests of detectability. Based on our findings, we place the attributes in the following order of concern: (1) Data source, (2) Fitzpatrick Skin Scale, (3) Ruler Presence, (4) Gentian marking presence (we skip localization, age and sex due to clinical relevance~\cite{carr2020epidemiology}). From Figure \ref{fig:natural_attributes} we see data source is both more detectable and higher in utility than other variables of interest, representing a potential shortcut. To the best of our knowledge, we are the first to document this concern, though recent independent work supports our result that differences between the sets are detectable \cite{Somafai2023DataDependence}. 

\section{Conclusions}
Our proposed method marks a positive step forward in anticipating and detecting unwanted bias in machine learning models. By focusing on dataset screening, we aim to prevent downstream models from inheriting biases already present and exploitable in the data. While our screening method naturally includes common auditing hypotheses (e.g., bias/fairness for vulnerable groups), it is capable of generating targeted hypotheses on a much broader set of attributes (e.g., sensor type, clinical collection site, imaging protocols, etc).  The ability to identify and investigate these hypotheses provides broad benefit for research, development, and regulatory efforts aimed at producing safe and reliable AI models.

\section{Acknowledgements}
This project was supported in part by an appointment to the Research Participation Program at the U.S. Food and Drug Administration administered by the Oak Ridge Institute for Science and Education through an interagency agreement between the U.S. Department of Energy and the U.S. Food and Drug Administration.

%
%
%
\bibliographystyle{splncs04}
\bibliography{egbib}

\end{document}


\title{Data-AUDIT: Supplementary Material}
\titlerunning{Data-AUDIT}
%
\author{}
\authorrunning{}
%
\institute{}
%

\section{Adjusted CMI Scaling}
\begin{figure}[]
    \centering
    \includegraphics[width=.9\textwidth]{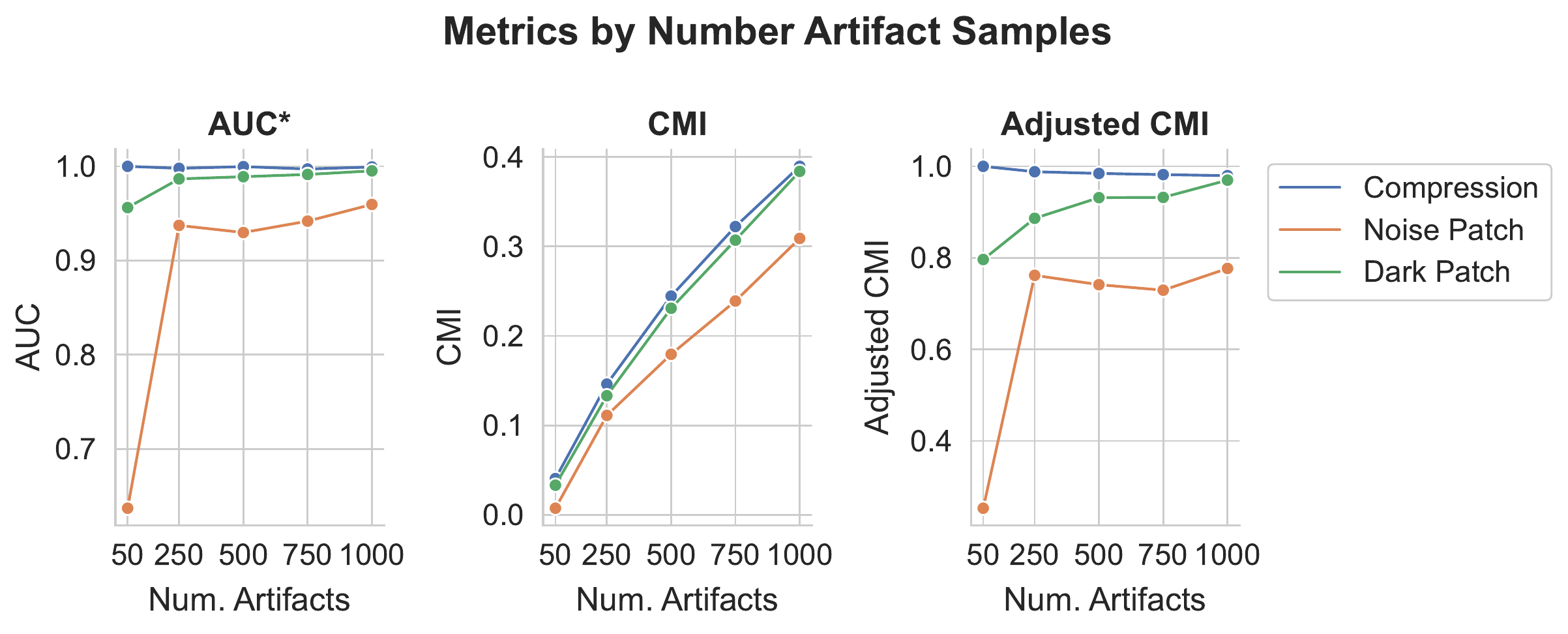}
    \caption{We compare the properties of Adjusted CMI, calculated as the expectation of Adjusted MI from Vinh et al. to raw CMI as the number of artifact examples increases. Left: AUC as calculated on rounded predictions. Middle: Raw CMI. Baseline of CMI increases with number of samples. Right: Adjustment for entropy and random chance yields a metric that is faithful to detection network performance and is stable across different numbers of examples.
    }
    \label{fig:scaling}
\end{figure}

\section{HAM10k Dataset}

\begin{table}[]
\caption{Distribution of malignant and benign lesions by data source. Because of variations in malignancy rates, identifying the source of an image provides valuable information about the likelihood of malignancy. Data sources include the Rosendahl practice in Queensland Australia (“Rosendahl”) and the ViDIR group in the Medical University of Vienna. ViDIR sources are subgrouped by image acquisition method, with further details in Tschandl et al. We use only one image per unique lesion for all tasks.}

\begin{tabular}{|l|lllll|}

\hline
\textbf{} & \multicolumn{4}{c|}{Data Source}                                                                                                                   & \multicolumn{1}{c|}{\multirow{}{Total}}\\ \cline{1-5} \cline{7-10}
          & \multicolumn{1}{l|}{Rosendahl} & \multicolumn{1}{l|}{ViDIR-Modern} & \multicolumn{1}{l|}{ViDIR-MoleMax} & \multicolumn{1}{l|}{Vienna Diapositives} & \multicolumn{1}{c|}{}                       \\ \hline
Malignant & \multicolumn{1}{l|}{41.6\%}       & \multicolumn{1}{l|}{27.7\%}          & \multicolumn{1}{l|}{0.3\%}            & \multicolumn{1}{l|}{17.3\%}                  & 15.7\%\\ \hline
Benign    & \multicolumn{1}{l|}{58.4\%}       & \multicolumn{1}{l|}{72.3\%}         & \multicolumn{1}{l|}{99.7\%}          & \multicolumn{1}{l|}{82.7\%}                 & 84.3\%\\ \hline
Number   & \multicolumn{1}{l|}{1549}      & \multicolumn{1}{l|}{1644}         & \multicolumn{1}{l|}{3916}          & \multicolumn{1}{l|}{278}                 & 7387\\ \hline
\end{tabular}
\end{table}

\section{Synthetic Datasets}
\begin{figure}[t]
    \centering
    \includegraphics[width=\linewidth]{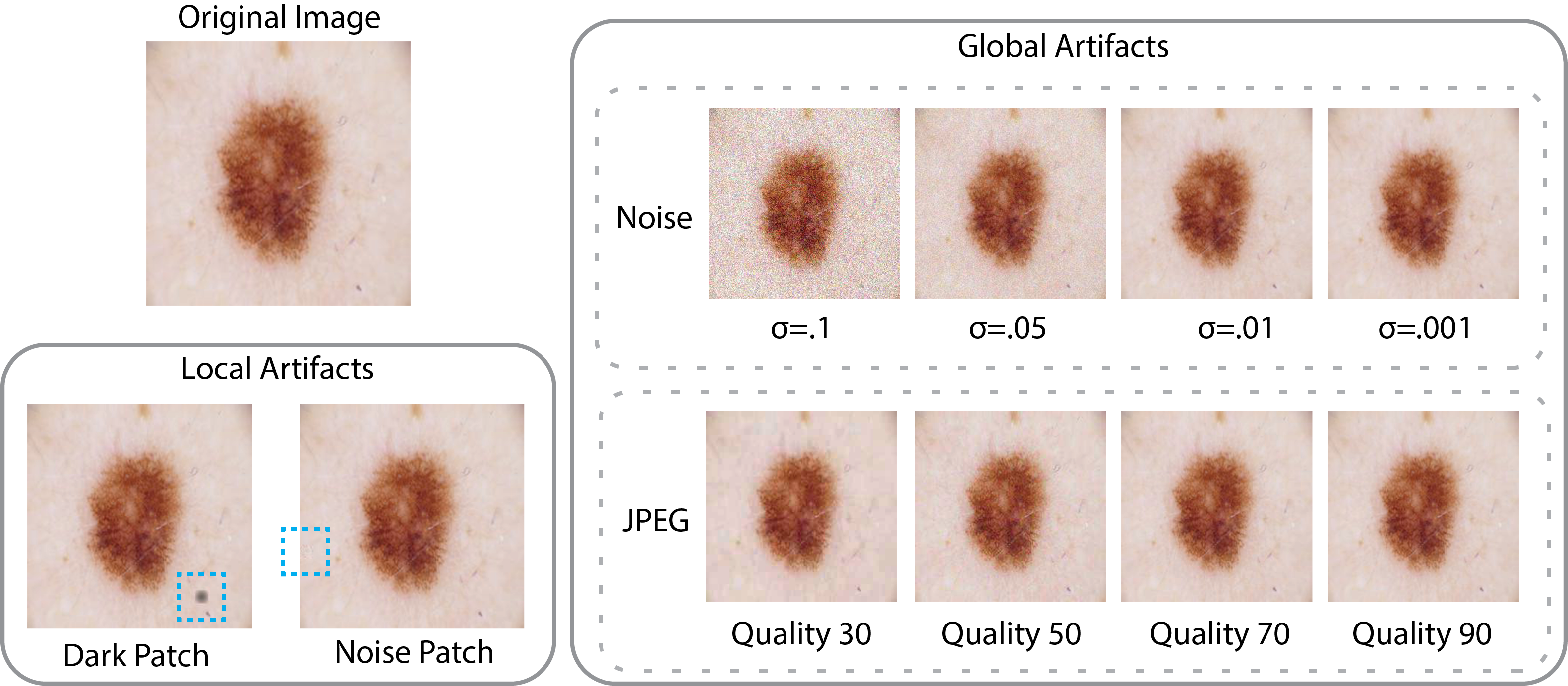}
    \caption{Examples of synthetic artifacts with varying image effects. Best viewed in color and at high magnification.
    }
    \label{fig:examples}
\end{figure}